\documentclass{article}
\pdfoutput=1
\usepackage{microtype}
\usepackage{graphicx}
\usepackage{subfig}
\usepackage{booktabs} 
\usepackage{amsmath}
\usepackage{tcolorbox}
\usepackage{amssymb}
\usepackage{amsthm}
\usepackage{lastpage}
\usepackage{fancyhdr}
\usepackage{accents}
\usepackage{titlesec}
\usepackage{listings}
\usepackage{enumitem}
\usepackage{footnote}
\setlist{nolistsep}
\usepackage{xcolor}
\usepackage{booktabs}

\usepackage{hyperref}
\usepackage{multicol}

\usepackage[accepted]{icml2021}
\icmltitlerunning{Syrupy Mouthfeel and Hints of Chocolate}

\begin{document}

\twocolumn[
\icmltitle{Syrupy Mouthfeel and Hints of Chocolate – Predicting Coffee Review Scores using Text Based Sentiment}



\icmlsetsymbol{equal}{*}

\begin{icmlauthorlist}
\icmlauthor{Christopher Lohse}{tcd}
\icmlauthor{Jeroen Lemsom}{tcd}
\icmlauthor{Athanasios Kalogiratos}{tcd}
\end{icmlauthorlist}
\icmlaffiliation{tcd}{Department of Computer Science and Statistics, Trinity College Dublin, Dublin, Ireland}
\icmlcorrespondingauthor{Christopher Lohse}{lohsec@tcd.ie}
\vskip 0.3in]



\printAffiliationsAndNotice{}  
\begin{abstract}
    This paper uses textual data contained in certified (q-graded) coffee reviews to predict corresponding scores on a scale from 0-100. By transforming this highly specialized and standardized textual data in a predictor space, we construct regression models which accurately capture the patterns in corresponding coffee bean scores.

\end{abstract}

\section{Motivation}

\begin{figure*}[h!]
    \centering
    \includegraphics[width = 0.8\textwidth]{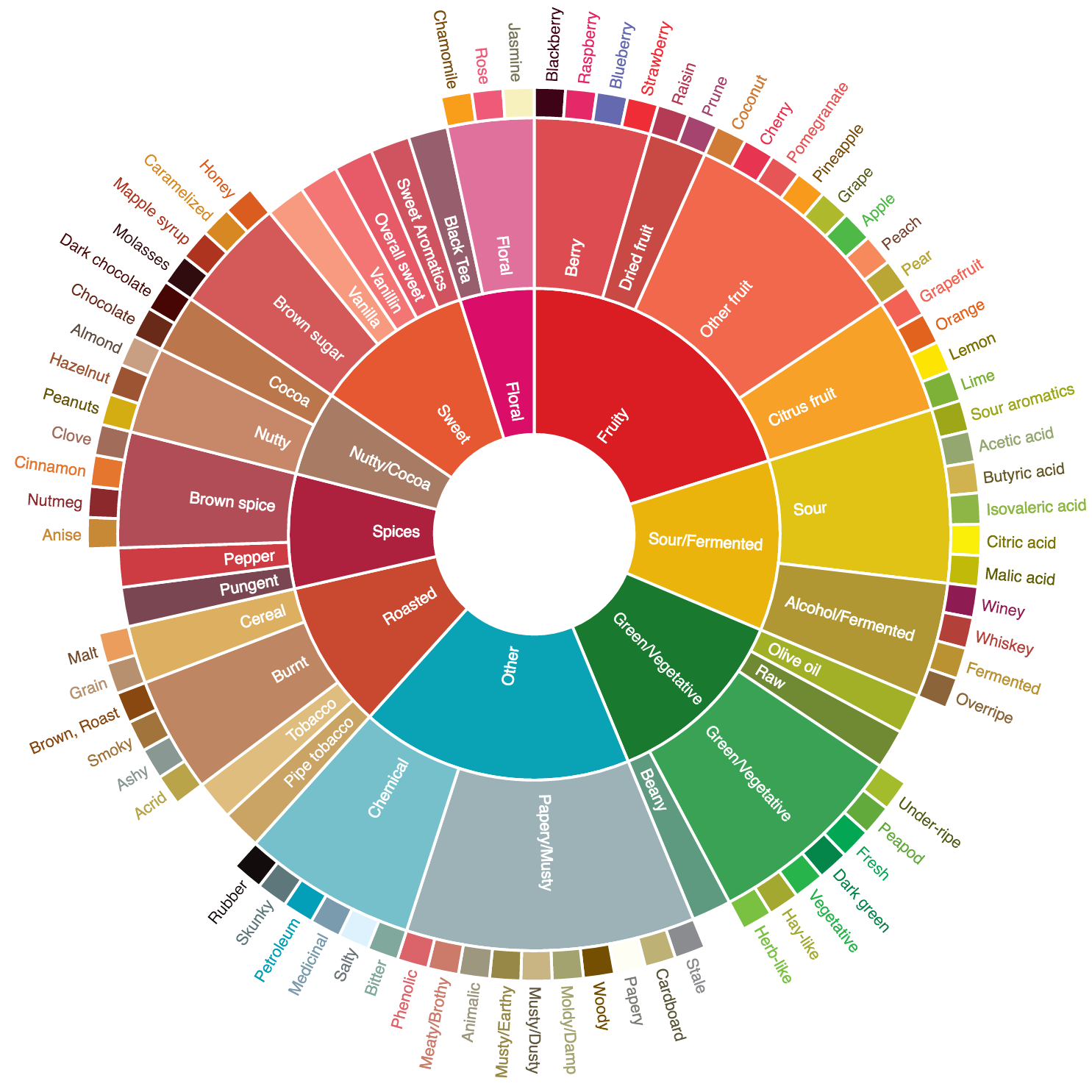}
    \caption[Speciality Coffee Taster Wheel]{Speciality Coffee Taster Wheel\footnotemark}
    \label{fig:wheel}
\end{figure*}

Coffee is one of the most traded goods worldwide – 2021 about 141 Million coffee bags 60 Kg each have been sold worldwide \cite{coffeexport}. To maintain a uniform standard in the assessment and therefore also the price of coffee, the Speciality Coffee Association introduced the so called q-grading system to rank coffee beans. The ranking is conducted by licensed coffee critiques—the so called q graders. To grade a coffee, q-graders taste the coffee at a tasting event. The coffees during these events are always brewed and served in the same way and scored by the q-graders based on the Speciality Coffee Taster Wheel (see Figure \ref{fig:wheel}) on a scale from 0-100 \cite{sca}. To support the score, the q-grader writes a short report with domain specific vocabulary to elaborate on their experience. Given the fact that the vocabulary is highly standardised, and that the coffees should be evaluated in a similar fashion, our hypothesis is that text based sentiment in q-graders reviews can be used to predict the corresponding scores. This would allow the Speciality Coffee Association to perform anomaly detection on coffee bean reviews and to verify whether the q-graders give scores which are in line with the standardised q system.




\section{Dataset} \label{sec:data}
For testing our hypothesis, we generate our own dataset by using employing web crawling techniques.
\subsection{Dataset Generation}
 The data set is based on roughly 6,000 coffee bean reviews published on the website \hyperlink{https://www.coffeereview.com/}{Coffereviews} going back to 1997. All of these reviews are scored with the q-grading scale \cite{coffeereview}.

\subsection{Dataset Cleaning}
To clean the dataset, we use the python library beautifulSoup\footnote{\url{https://www.crummy.com/software/BeautifulSoup/}} to parse the cleartext from the roughly one million lines of HTML texts of the pages which have been  downloaded using the python requests\footnote{\url{https://requests.readthedocs.io/en/latest/}} library before.

After cleaning the dataset, we have two features in the dataset;  one feature $i$ which is the text of the review and the corresponding score $y_i$ which is the target variable for the prediction. During the cleaning process, we had to drop $2$ points from the dataset since they did not have a score.
Additionally, we had to remove roughly 1000 feature, which had an empty text – resulting in 4985 unique features in our dataset.

\begin{figure}[!h]
    \centering
    \includegraphics[width = 0.45\textwidth]{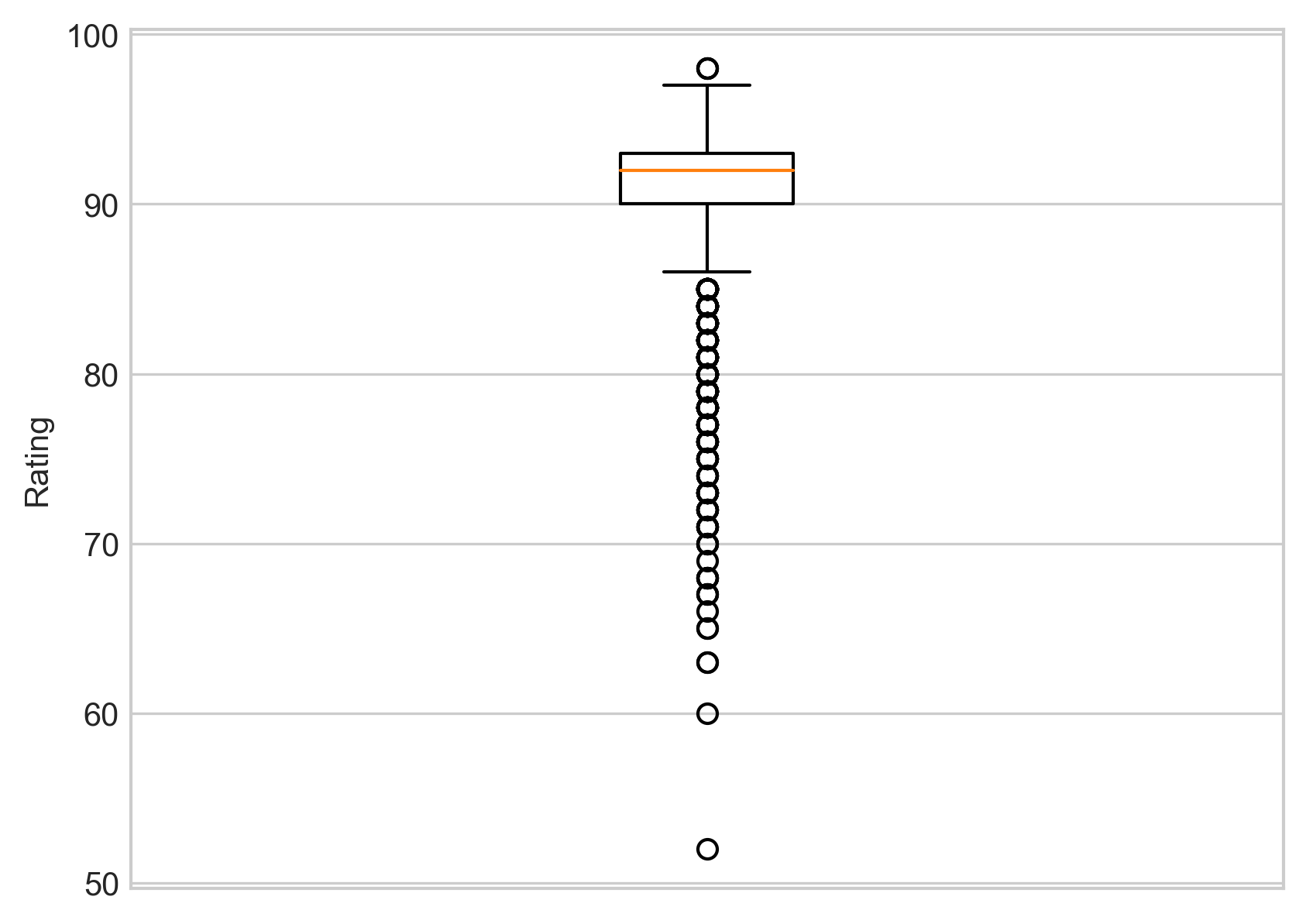}
    \caption{Boxplot of Coffee Bean Scores}
    \label{fig:boxplot}
\end{figure}

Figure \ref{fig:boxplot} shows that most of the scores in our dataset are between 86 and 97. This implies that most of the reviewed coffee beans receive the qualification `speciality coffee', which requires a score of 85 out of 100 points \cite{poltronieri2016challenges}.

\subsection{Feature Engineering}
To capture the text more accurately, we decided to remove t
stop words from the reviews; stop words are words which occur very often in every text, e.g. and/or and therefore do not have a high relevance of a text's sentiment. For the stop words removal, we used the standard English stop word set from the python library NLTK.\footnote{\url{https://www.nltk.org/}}
As input feature for our model we take unigrams of words – which are just single words in the text and bigrams which are the combination of two words occurring in the texts. This leads to a significant larger number of features for models which have bigrams as their input. (5111 features for unigram and 53341 for bigrams)

\subsection{Data Analysis}

\begin{figure}[h!]
    \centering
    \includegraphics[width = 0.5\textwidth]{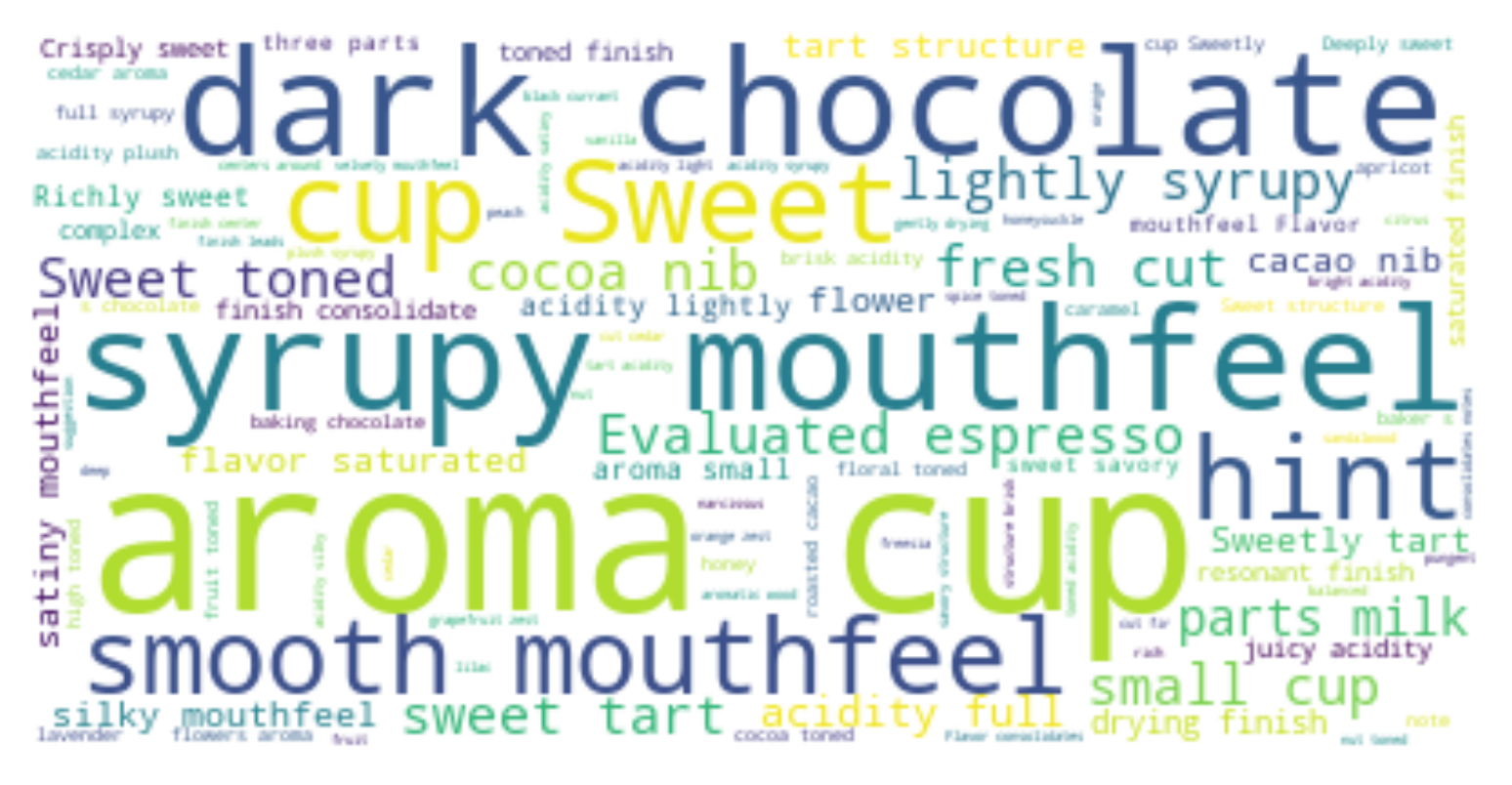}
    \caption{Wordcloud with the most frequent bigrams and unigrams}
    \label{fig:cloud}
\end{figure}
Figure \ref{fig:cloud} shows the most frequent unigrams and bigrams contained in the text reviews after the stopword removal – this illustrates the highly specific domain vocabulary with terms such as `hint', `note flower' and `syrupy mouthfeel'.

\section{Methodology}\label{chap:meth}

We compare a selection of different machine learning models to predict coffee bean scores $y_{i}$ which range between 0 and 100 for coffee bean $i$, where $i \in \{1,...,N\}$ based on the sentiment in written reviews. The performance of the model specifications in this section will be compared against a naive predictor, which always predicts the average score.

\subsection{Bag of Words Regression} \label{sec:bagofwords}
Our first model is the bag of words regression which can be represented as
\begin{equation} \label{eq:bagofwordsregression}
    y_{i} = \alpha + \beta_{1} x_{i,1} + \beta_{2} x_{i,2} + ... + \beta_{m} x_{i,m},
\end{equation}
where $y_{i}$ is the numerical score corresponding to review $i$. Here, $x_{i,j}$ equals the number of times term $j$ appears in review $i$. Moreover, $\alpha$ defines the intercept and $\beta_{j}$ reflects the sentiment corresponding to term $j$. If $\beta_{j}$ is positive, an occurrence of term $j$ in a review will lead to an increase of the score. On the other hand, when $\beta_{j}$ is negative, the term carries a negative sentiment.


\subsection{TF-IDF Regression}
\subsubsection{Without Regularisation}
In contrast to the bag of words regression discussed in Section \ref{sec:bagofwords}, the TF-IDF regression uses transformed predictors $z_{i,j}$. The Term Frequency (TF) of term $j$ in review $i$ scales the occurrences of term $i$ by the total number of terms counted in review $j$. This can be represented as
\begin{equation}
    TF_{i,j} = \frac{\sum_{j} x_{i,j}}{\sum_{i} \sum_{j} x_{i,j}}.
\end{equation}
The Inverse Document Frequency (IDF) quantifies the uniqueness of term $j$ across all documents $i \in \{1, ..., N\}$. The IDF is defined as
\begin{equation}
    IDF_{i,j} = log \left( \frac{N}{1 + \sum_{j} I(i,j)} \right),
\end{equation}
where $N$ denotes the total number of reviews, while indicator $I(i,j) = 1$ if term $i$ appears in document $j$ and $I(i,j) = 0$ otherwise. The transformed predictor $z_{i,j}$ is then constructed as
\begin{equation}
    z_{i,j} = TFIDF_{i,j} = TF_{i,j} IDF_{i,j}.
\end{equation}
Then the TF-IDF regression is specified as
\begin{equation}
\label{eq:tfidfregression}
    y_{i} = \alpha + \beta_{1} z_{i,1} + \beta_{2} z_{i,2} + ... + \beta_{m} z_{i,m}.
\end{equation}

\subsubsection{With Ridge Regularisation} \label{sec:tfidfridge}
The coefficients $\boldsymbol{\beta} = [\beta_{1}, .. ,\beta_{m}]$ in Equation \ref{eq:tfidfregression} is calculated by minimizing the mean squared error loss cost function
\begin{equation} \label{eq:optimization}
    \frac{1}{p} \sum_{i=1}^{p} (\boldsymbol{\beta}^{\boldsymbol{T}} \boldsymbol{z_{i}} - y_{i})^{2}.
\end{equation}
Note that $\boldsymbol{\beta^{T}}$ is a (1 x m) dimensional vector, while $\boldsymbol{{z}_{i}}$ is (m x 1) dimensional\footnote{Similarly the coefficients in the bag of words regression in Equation \ref{eq:bagofwordsregression} in Section \ref{sec:bagofwords} are also calculated by minimizing the mean squared error loss. In that case, one needs to substitute $\boldsymbol{z_{i}}$ in Equation \ref{eq:optimization} by $\boldsymbol{x_{i}}$}. A number of p (p < N) training observations is used to train the model and obtain the coefficients in $\boldsymbol{\beta}$. When including a ridge penalty term, the loss function becomes
\begin{equation} \label{eq:optimizationridge}
    \frac{1}{p} \sum_{i=1}^{p} (\boldsymbol{\beta}^{\boldsymbol{T}} \boldsymbol{z_{i}} - y_{i})^{2} + \frac{1}{2C} \sum_{j=1}^{m} \beta_{j}^{2}.
\end{equation}
Adding a ridge penalty term in our optimization could prevent overfitting in case the number of predictors $m$ is very large. An appropriate hyperparameter $C$ should be selected based on k-fold cross validation.




\subsection{K-NN Regression} \label{sec:knn}
In the K-NN Regression, $\hat{y}$ corresponding to test observation $(y,z)$ will be based on the average score of the $k$ closest points measure by the Euclidean distance measure in a $m$ dimensional space (which is the total number of terms in our training set). For the K-NN regression, independent variables $z_{j}$ are equal to the independent variables in the TF-IDF regression. The Euclidean distance from $(y,z)$ to a training point $(y^{(i)},z^{(i)})$ is
\begin{equation}
    d(z^{(i)},z) = \sqrt{\sum_{j=1}^{m} (z_{j}^{(i)} - z_{j})^{2}}
\end{equation}
If $N_{k}$ contains the $k$ observations closest to $z$, then
\begin{equation}
    \hat{y} = \frac{\sum_{i \in N_{k}} y^{(i)}}{k}.
\end{equation}


\section{Experimental Setup}
\subsection{K-Fold Cross Validation} \label{sec:k-fold}
The TF-IDF Ridge regression in Section \ref{sec:tfidfridge} and the K-NN Regression in Section \ref{sec:knn} both contain a hyperparameter. These hyperparameters will be optimized to obtain the best value; the best model will then be considered in the benchmarking. These hyperparameters are optimized by performing k-fold cross validation on our training set with $kf = 5$ folds. This means that the training data is split into $kf$ folds; then $kf$ models are trained — each time with a different fold left out.


\subsection{Performance Measures}

The performance of a model is quantified using the Mean Squared Error (MSE) measure, which can be represented as
\begin{equation}
    MSE = \sum_{i=1}^{n}(y_i-\hat{y}_{i})^2,
\end{equation}
where $y_{i}$ is the actual score and $\hat{y}_{i}$ is the predicted score for review $i$. A characteristic of the Mean Squared Error (MSE) over other performance measures is that it penalises larger mistakes more heavily. The MSE will be used to decide upon the hyperparameter values of the TF-IDF Ridge Regression and K-NN regressions in our K-Fold cross validation setup.

When comparing our full range of models, we additionally compute the MAE (Mean Absolute Error) measures
\begin{equation}
    MAE = \sum_{i=1}^{n}|y_{i}-\hat{y}_{i}|.
\end{equation}
The Mean Squared Error is used to select the optimal hyperparameters when performing k-fold cross validation in Section \ref{sec:k-fold} and the compare the out-of-sample performance of our various model specifications.


\section{Evaluation}

\subsection{Hyperparameter Optimization}

The optimal hyperparameter values $k$ for the unigram and bigram K-NN regressions are both 11, while the optimized value $C$ for the unigram and bigram Ridge TF-IDF Regressions are equal to 1 and 10 respectively. The grids of values which are used for the in-sample optimization of the hyperparameters are shown in the Appendix and are accompanied by Figure \ref{fig:hypeeerrrrrr} which shows a plot containing the MSE values for different hyperparameter values. It becomes apparent in the plot that both of the ridge regression models have a lower standard deviation; therefore, one conclude that these models yield into more stable results. Moreover, the unigram ridge model has again a lower standard deviation than the bigram model. We note that the unigram model of the Ridge regression is better with higher penalisation than the bigram model.

\subsection{Performance}

Table \ref{tab:results} shows the out-of-sample performance of the models discussed in Section \ref{chap:meth}.

\begin{table}[h]
\caption{Out-of-sample performance of all models. Best results are printed in 
\textbf{bold}}
\label{tab:results}
\centering
\footnotesize
\begin{tabular}{lcc}
\hline
\multicolumn{1}{c}{Model} & MSE   & MAE   \\ \hline
Naive Mean                      & 6.624 & 1.644 \\
BoW Regression Unigram    & 3.794 & 1.277 \\
BoW Regression Bigram     & 2.980 & 1.077 \\
TF-IDF Regression Unigram & 2.990 & 1.192 \\
TF-IDF Regression Bigram  & 2.581 & 1.046 \\
Ridge TF-IDF Unigram      & \textbf{1.508} & \textbf{0.864} \\
Ridge TF-IDF Bigram       & 2.582 & 1.034 \\
KNN Regression Unigram    & 2.277 & 1.038 \\
KNN Regression Bigram     & 3.472 & 1.111 \\ \hline
\end{tabular}
\end{table}

The K-NN Regressions, which take the average score of the closest 11 observations in the training set, improves upon the Naive Mean benchmark which always predicts the mean score of the scores in the training set. Furthermore, Table \ref{tab:results} shows that also our other regression specifications achieve a better out-of-sample performance than our benchmark model. Based on both the MSE and MAE performance measures, the Ridge TF-IDF regression using unigrams turns out to be most effective in predicting coffee scores based on written reviews. This finding suggests that it is beneficial to shrink the weights corresponding to non-significant terms using the ridge penalty term. Examples of score predictions based on textual reviews from the Ridge TF-IDF Regression Unigram model are illustrated in Table \ref{tab:example} in the Appendix. The Table contains 10 reviews randomly drawn form the test set and show that the predictions are relatively accurate.

\subsection{Interpretation}

\begin{figure}[h!]
    \centering
    \includegraphics[width = 0.5\textwidth]{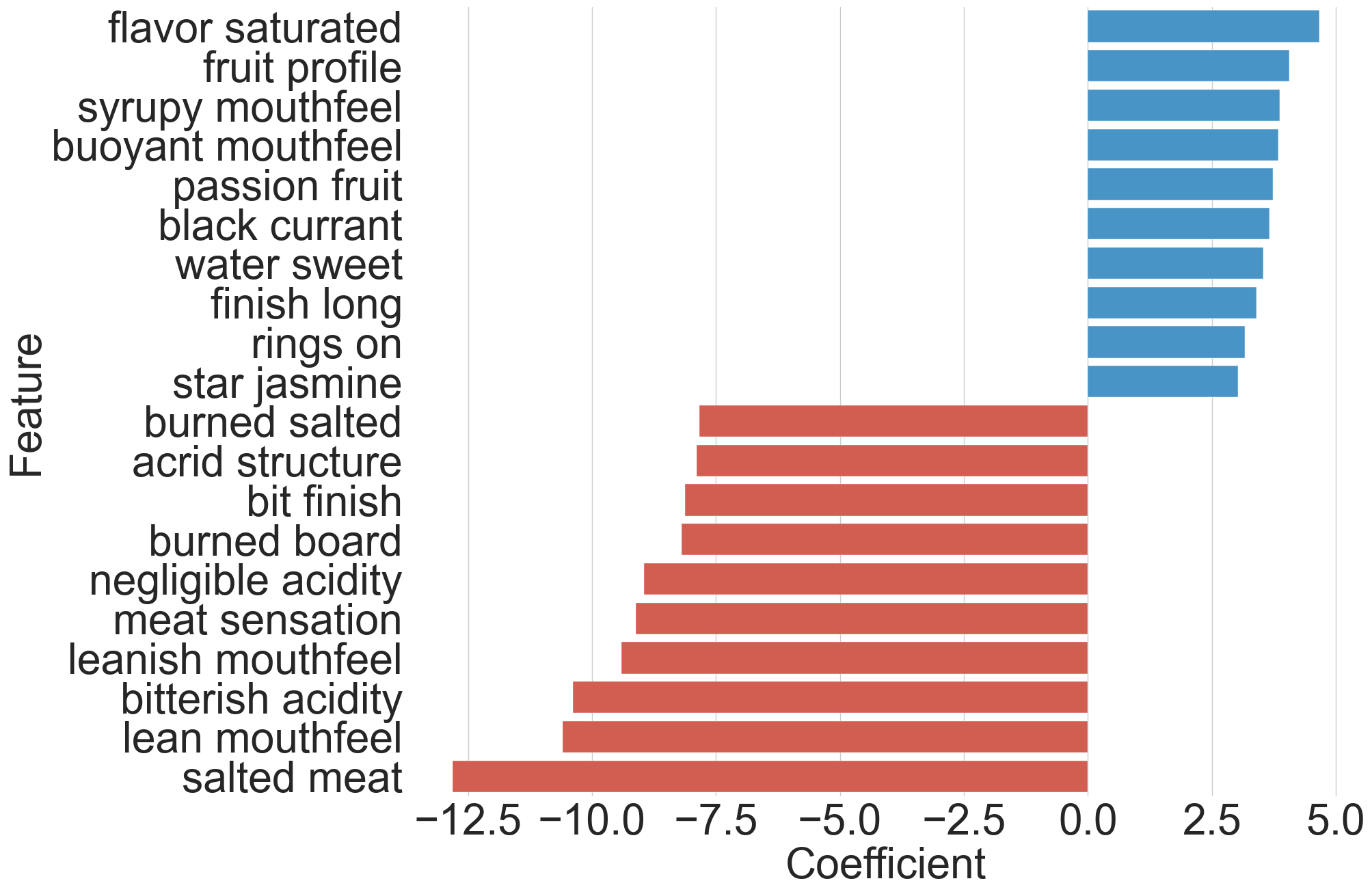}
    \caption{Bigrams with Strongest Positive and Negative Sentiment}
    \label{fig:interpret}
\end{figure}

Figure \ref{fig:interpret} presents the bigrams with the most positive and negative sentiments in the Ridge TF-IDF bigram regression.

The results indicate that a q-graders experience of the `mouthfeel' could strongly affect the resulting score. In particular, `syrupy' or `buoyant' mouthfeels tend to describe higher-ranked coffees, while the bigram `leanish mouthfeel' carries a strong negative sentiment. Besides, q-graders appreciate coffees with a `long finish' and fruity notes such as black currant and passion fruit carry a strong positive sentiment. On the other hand, descriptions of a `salty' or `meaty' flavor decrease the score of a coffee bean. In contrary to a `long finish', a `bit finish', which is used to describe a `short finish', carries negative sentiment.

\section{Conclusion}
By creating a wordcloud of the terms in the reviews and by highlighting the terms with the strongest positive and negative sentiments in a penalized TF-IDF regession, we show that the vocabulary used in the coffee reviews is highly specialized. 

Our findings indicate that textual data from q-graded coffee bean reviews has strong predictive value when forecasting corresponding scores on a scale from 0-100. In particular, we find that both a bag of word regression and a regression with TF-IDF adjusted predictors offer significant advantages over a naive mean benchmark predictor. Additionally, we find that adding an (optimized) ridge penalty term to the TF-IDF regression leads to an improved  out-of-sample performance.

In conclusion, the unigrams and bigrams extracted from the textual coffee reviews show to be relevant independent variables when predicting coffee review scores in a regression context, due to the highly standardized terminology used by certified q-graders. We encourage further research to assess the predictive performance of our models in combination with a more thorough pre-processing of the textual data, including techniques such as lemmatization, tokenization and stemming.

\bibliographystyle{icml2021}
\bibliography{main}

\clearpage
\onecolumn
\section*{Appendix}
\subsection*{Flavour Wheel}
\footnotetext{Retrieved from \url{https://notbadcoffee.com/flavor-wheel-en/}}
\clearpage

\subsection*{Hyperparameter Optimization}
For hyperparameter $C$ in the ridge TF-IDF regression, we consider the grid \newline $[0.0001, 0.001, 0.01, 0.1, 1, 10, 20]$. For $k$ in the K-NN regression, we consider the grid \newline $[1, 11, 21, 51, 101, 201]$\footnote{For the K-NN regression it is not required to consider 
 exclusively odd values in the grid. This would only be necessary for classification problems}.

\begin{figure*}[h!]
    \centering
\subfloat[Ridge Unigram]{\includegraphics[width=0.49\textwidth]{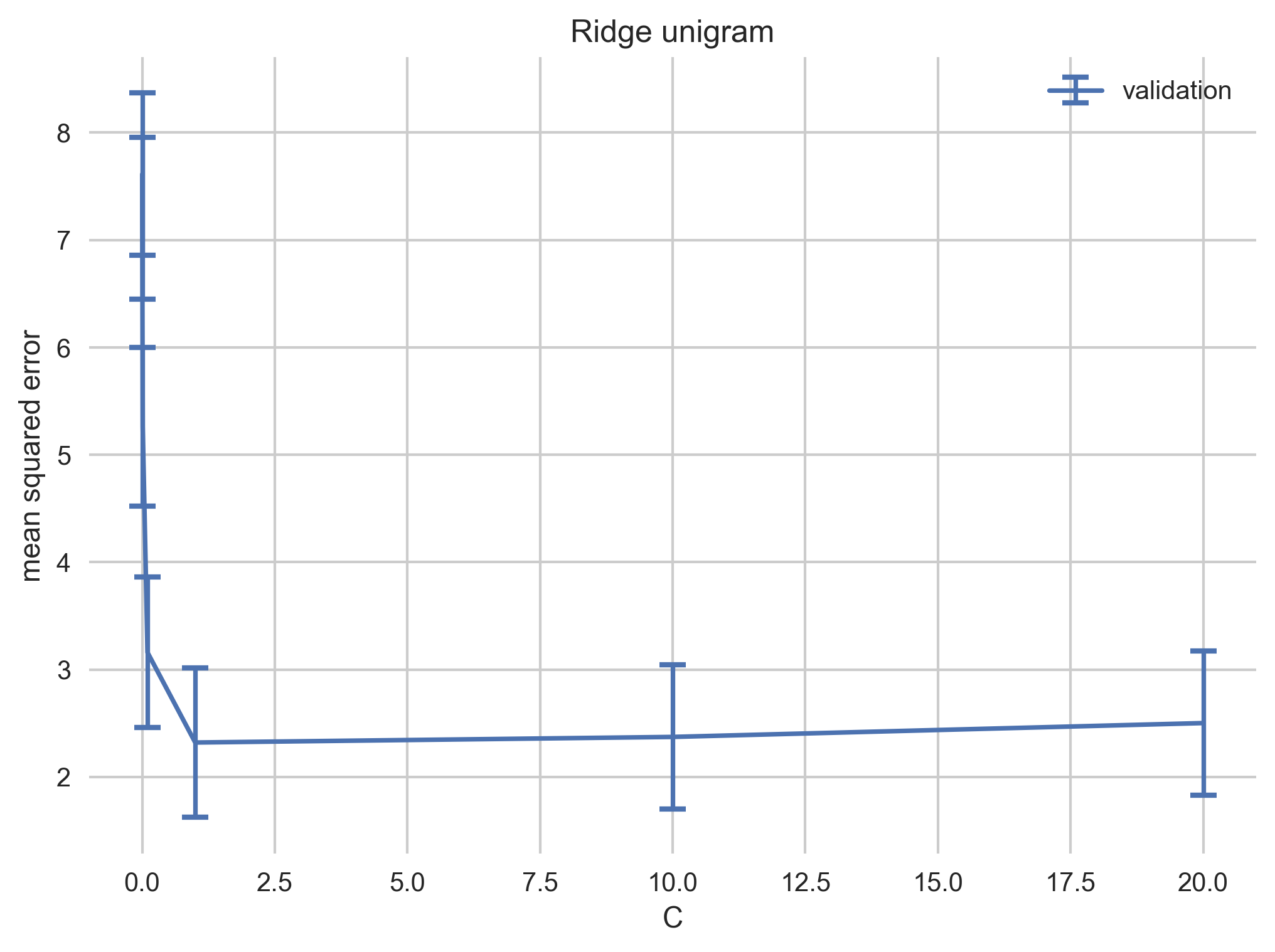}
\label{fig:r_ug}}
\subfloat[Ridge Bigram]{
    \includegraphics[width=0.49\textwidth]{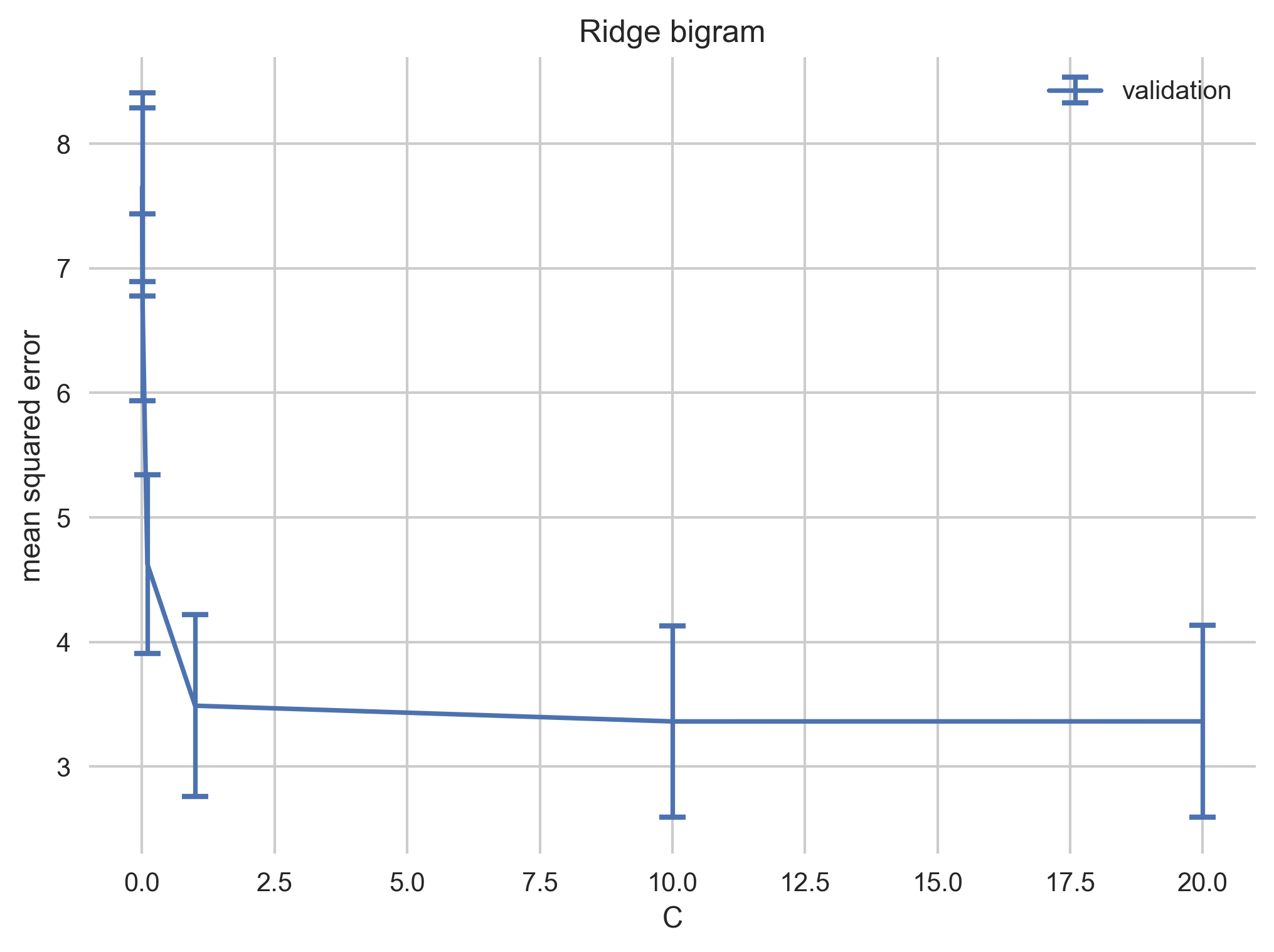}
    \label{fig:r_bg}}

\subfloat[KNN Unigram]{
    \includegraphics[width=0.49\textwidth]{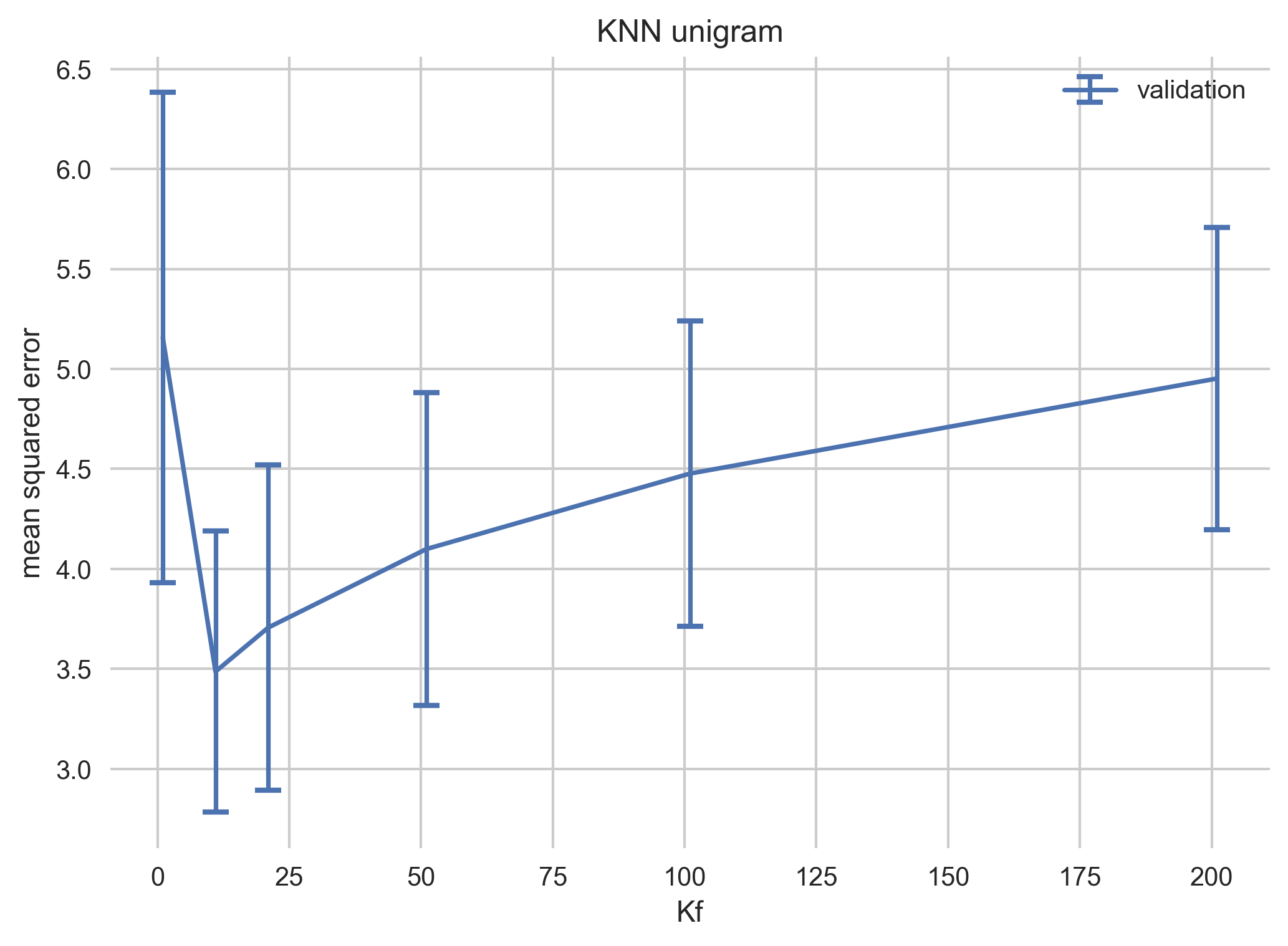}
    \label{fig:knn_bg}}
\subfloat[KNN Bigram]{
    \includegraphics[width=0.49\textwidth]{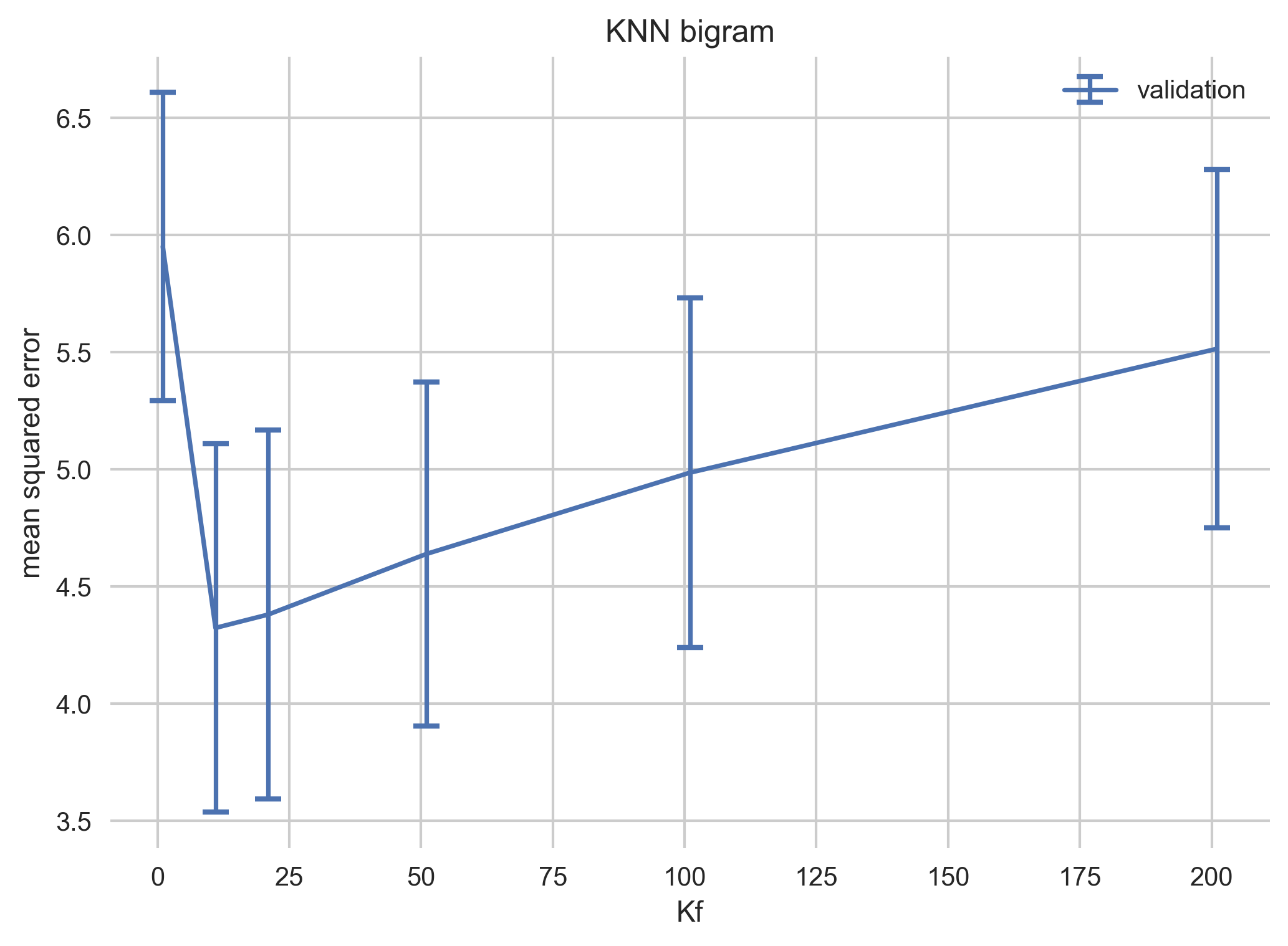}   \label{fig:knn_bg}}
\caption{Hyperparameter Optimization }  \label{fig:hypeeerrrrrr}
\end{figure*}
\clearpage
\subsection*{Examples on Test Set}
\begin{table}[h!]
\caption{10 Example Predictions: numerical predictions are rounded to the nearest integer value}
\label{tab:example}
\resizebox{\textwidth}{!}{%
\begin{tabular}{@{}lll@{}}
\toprule
text                                                                                                                                                                                                                                                                                                                             &true & pred \\ \midrule
\begin{tabular}[c]{@{}l@{}}Luscious, deeply sweet, sumptuous. Blackberry, lavender, marjoram, roasted cacao nib, grapefruit zest aroma cup.\\ Rich, juicy acidity; smooth, syrupy mouthfeel. The long, flavor-saturated finish zesty sweet, offering surprise appearance\\ tropical fruit notes (coconut) resonate long while.\end{tabular} & 97.0 & 95.0 \\ \midrule
\begin{tabular}[c]{@{}l@{}}Delicate, intricate, expressive. Almond butter, plumeria, roasted cacao nib, myrrh, loquat aroma cup.\\ Lively sweet-tart acidity; silky mouthfeel. The sweet-toned, resonant finish centers around plumeria\\ cocoa, suggestions myrrh short finish loquat long.\end{tabular}                                   & 94.0 & 94.0 \\ \midrule
\begin{tabular}[c]{@{}l@{}}Crisply sweet, cocoa-toned. Baking chocolate, hazelnut, date, cedar, molasses aroma cup.\\ Sweet structure gentle, round acidity; full, velvety mouthfeel.\\ The finish consolidates baking chocolate, hazelnut cedar.\end{tabular}                                                                              & 92.0 & 91.0 \\ \midrule
\begin{tabular}[c]{@{}l@{}}Delicate yet lush, gently bright. \\ Dried peach, ripe orange, candied walnut, hints fresh-cut cedar cinnamon aroma cup.\\ Roundly balanced acidity; satiny, light-footed mouthfeel. \\ Walnut, cedar orange particular carry crisp sweet finish.\end{tabular}                                                   & 93.0 & 93.0 \\ \midrule
\begin{tabular}[c]{@{}l@{}}High-toned, crisply sweet. Date, cashew butter, magnolia, grapefruit zest, cedar aroma cup.\\ Sweetly tart structure brisk acidity; full, velvety mouthfeel.\\ The resonant finish leads notes date magnolia short, rounding cashew butter grapefruit zest long.\end{tabular}                                    & 92.0 & 92.0 \\ \midrule
\begin{tabular}[c]{@{}l@{}}Suavely bright, fragrant, deep. Flowers, pear- peach-like fruit, honey, pistachio-like nut aroma cup. \\ Rich acidity; full, syrupy mouthfeel. Flavor consolidates resonant finish\end{tabular}                                                                                                                  & 94.0 & 93.0 \\ \midrule
\begin{tabular}[c]{@{}l@{}}Delicately complex: black tea, lemon-lime, spicy rose, moist pipe tobacco.\\ Gently bright acidity; lean silky mouthfeel. Flavor consolidates cocoa crisp finish.\end{tabular}                                                                                                                                   & 91.0 & 91.0 \\ \midrule
\begin{tabular}[c]{@{}l@{}}Sweetly savory, vibrant, deep. Meyer lemon zest, dark chocolate, tamarind, sandalwood, aromatic orchid aroma cup.\\ Sweet-tart structure bright, juicy acidity; satiny-smooth mouthfeel.\\  Flavor-saturated-finish leads notes tamarind lemon zest, rounding deep, dark chocolate notes long.\end{tabular}      & 93.0 & 95.0 \\ \midrule
\begin{tabular}[c]{@{}l@{}}Richly sweet, deeply spice-toned. Toffee, plum blossom, tangerine zest, almond butter, sandalwood aroma cup.\\ Juicy-sweet structure high-toned, balanced acidity; syrupy-smooth mouthfeel.\\ The finish centers around notes almond, tangerine zest sandalwood.\end{tabular}                                    & 94.0 & 94.0 \\ \midrule
\begin{tabular}[c]{@{}l@{}}Very sweet, gently bright, delicately fruit-toned. Lush flowers, cidery apple tart, raspberry-like fruit aroma cup.\\ Softly bright acidity; silky lightly syrupy mouthfeel. Very sweet finish.\end{tabular}                                                                                                     & 93.0 & 93.0 \\ \bottomrule
\end{tabular}%
}
\end{table}


\end{document}